\documentclass[sigconf,screen]{acmart}

\usepackage{booktabs}
\usepackage{amsfonts}
\usepackage{graphicx}
\usepackage{hyperref}
\usepackage{textcomp}
\usepackage{listings}
\usepackage{adjustbox}
\usepackage{xspace}    
\usepackage{multirow}
\usepackage{multicol}
\usepackage{xcolor} 
\usepackage{amsmath}
\usepackage{tcolorbox}
\usepackage{algorithm}
\usepackage{algorithmic}
\usepackage{listings}

\def \tool{\texttt{DyCL}\xspace}

\def \tvm{\texttt{TVM}\xspace}
\def \onnx{\texttt{OnnxRuntime}\xspace}

\newcommand{\eg}{{\it e.g.,}\xspace}

\newcommand{\etc}{{\it etc.}\xspace}
\newcommand{\ie}{{\it i.e.,}\xspace}

\newcommand\figref[1]{Fig.~\ref{#1}}

\newcommand\tabref[1]{Table~\ref{#1}}

\newcommand\secref[1]{Section~\ref{#1}}
\newcommand\equref[1]{Eq.(\ref{#1})}

\usepackage{wasysym}

\definecolor{codegreen}{rgb}{0,0.6,0}
\definecolor{codegray}{rgb}{0.5,0.5,0.5}
\definecolor{codepurple}{rgb}{0.58,0,0.82}
\definecolor{backcolour}{rgb}{0.95,0.95,0.92}

\lstdefinestyle{mystyle}{
  backgroundcolor=\color{backcolour},   commentstyle=\color{codegreen},
  keywordstyle=\color{magenta},
  numberstyle=\tiny\color{codegray},
  stringstyle=\color{codepurple},
  basicstyle=\ttfamily\footnotesize,
  breakatwhitespace=false,         
  breaklines=true,                 
  captionpos=b,                    
  keepspaces=true,                 
  numbers=right,                    
  numbersep=-5pt,                  
  showspaces=false,                
  showstringspaces=false,
  showtabs=false,                  
  tabsize=1,
}
\newcommand{\fakeparagraph}[1]{\vspace{1mm}\noindent\textbf{#1.}}
\usepackage{hyperref}
\usepackage{microtype}

\setcopyright{rightsretained}
\acmPrice{}
\acmDOI{10.1145/3597926.3598082}
\acmYear{2023}
\copyrightyear{2023}
\acmSubmissionID{issta23main-p158-p}
\acmISBN{979-8-4007-0221-1/23/07}
\acmConference[ISSTA '23]{Proceedings of the 32nd ACM SIGSOFT International Symposium on Software Testing and Analysis}{July 17--21, 2023}{Seattle, WA, USA}
\acmBooktitle{Proceedings of the 32nd ACM SIGSOFT International Symposium on Software Testing and Analysis (ISSTA '23), July 17--21, 2023, Seattle, WA, USA}
\received{2023-02-16}
\received[accepted]{2023-05-03}

\ccsdesc[500]{Software and its engineering~Compilers}
\ccsdesc[500]{Computing methodologies~Machine learning}

\keywords{Dynamic Neural Networks, Deep Learning Compilers}

\begin{document}

\title{DyCL: Dynamic Neural Network Compilation Via Program Rewriting and Graph Optimization}


\author{Simin Chen}
\email{simin.chen@UTDallas.edu}
\affiliation{%
  \institution{UT Dallas}
  \city{Dallas}
  \country{USA}
}

\author{Shiyi Wei}
\email{swei@utdallas.edu}
\affiliation{%
  \institution{UT Dallas}
  \city{Dallas}
  \country{USA}
}

\author{Cong Liu}
\email{congl@ucr.edu}
\affiliation{%
  \institution{UC Riverside}
  \city{Riverside}
  \country{USA}
}

\author{Wei Yang}
\email{wei.yang@utdallas.edu}
\affiliation{%
  \institution{UT Dallas}
  \city{Dallas}
  \country{USA}
}

\begin{abstract}

The deep learning (DL) compiler serves as a vital infrastructure component to enable the deployment of deep neural networks on diverse hardware platforms such as mobile devices and Raspberry Pi. 
DL compiler's primary function is to translate DNN programs written in high-level DL frameworks such as PyTorch and TensorFlow into portable executables. These executables can then be flexibly executed by the deployed host programs.
However, existing DL compilers rely on a tracing mechanism, which involves feeding a runtime input to a neural network program and tracing the program execution paths to generate the computational graph necessary for compilation. 
Unfortunately, this mechanism falls short when dealing with modern dynamic neural networks (DyNNs) that possess varying computational graphs depending on the inputs. Consequently, conventional DL compilers struggle to accurately compile DyNNs into executable code.
To address this limitation, we propose \tool, a general approach that enables any existing DL compiler to successfully compile DyNNs. 
\tool tackles the dynamic nature of DyNNs by introducing a compilation mechanism that redistributes the control and data flow of the original DNN programs during the compilation process.
Specifically, \tool develops program analysis and program transformation techniques to convert a dynamic neural network into multiple sub-neural networks. Each sub-neural network is devoid of conditional statements and is compiled independently. Furthermore, \tool synthesizes a host module that models the control flow of the DyNNs and facilitates the invocation of the sub-neural networks.
Our evaluation demonstrates the effectiveness of \tool, achieving a 100\% success rate in compiling all dynamic neural networks. Moreover, the compiled executables generated by \tool exhibit significantly improved performance, running between $1.12\times$  and $20.21\times$ faster than the original DyNNs executed on general-purpose DL frameworks.

\end{abstract}

\maketitle

\section{Introduction}

With the growing popularity of deep learning(DL)-based applications, optimizing, executing, and deploying these applications becomes critical.
DL compilers~\cite{tvm, glow, onnxruntime, tensorflowLite, reverse2022, zheng2020ansor, zheng2022dietcode, fegade2021cortex, fang2021eto, lyubomirsky2022compiler} are fundamental infrastructures for achieving these goals, enabling DL application deployment on various hardware devices.
DL compilers translate DL models written in high-level DL frameworks (\eg \texttt{PyTorch} \cite{paszke2019pytorch} and \texttt{TensorFlow} \cite{abadi2016tensorflow}) into optimized and portable executables.

Over the past few years, significant advancements have been made in the development of DL compilers, aiming to streamline the deployment of neural networks on diverse hardware platforms \cite{chen2018tvm, glow,cyphers2018intel,vanholder2016efficient}.
DL compilers offer two key advantages. First, they enable the compiled DNN model to function as a executable program, eliminating the need for model developers to install resource-intensive DL frameworks on target platforms to parse and execute the DNN models.
Second, DL compilers optimize the inference time overhead of a given DNN model, making it suitable for real-time applications on resource-constrained platforms such as mobile devices.

However, existing DL compilers heavily rely on the \emph{tracing mechanism} \cite{chen2018tvm, onnxruntime}, which requires providing a runtime input to a neural network program and tracing its execution path to generate the necessary computational graph for compilation.
Unfortunately, these tracing mechanisms implicitly assume that a DNN model can be abstracted as a ``static'' computational graph, with a fixed execution path for computation.
This assumption, however, does not hold for modern dynamic neural networks (DyNNs), where the execution path is determined by individual inputs and varies with each invocation \cite{adnn1, adnn2, adnn3, adnn4, adnn5, adnn6}.
For instance, an encoder-decoder model used in neural machine translation \cite{chen2022nmtsloth} may require invoking the underlying decoder multiple times to produce translation outputs, without specifying the exact number of invocations.

To understand the limitations of the tracing mechanism employed by existing DL compilers when it comes to compiling dynamic neural networks, we conducted an empirical study utilizing two widely-used DL compilers (\ie \texttt{TVM} \cite{chen2018tvm} and \texttt{OnnxRuntime} \cite{onnxruntime}) to compile four types of DyNNs. 
The results revealed a discrepancy between the outputs produced by running the compiled executables and the original DNN models within DL frameworks. This discrepancy clearly indicates that the DL compilers fail to accurately compile DyNNs.

To overcome the limitations of existing DL compilers, we introduce an automatic tool, \tool, which assists developers in accurately compiling DyNNs automatically.
Our primary objective is to enable the flexible adaptation of optimizations found in current DL compilers while ensuring the correct handling of the inherent dynamism present in DyNNs.
The design of \tool is driven by two key observations. First, we identify that the main source of DL compilers' inability to compile DyNNs lies in the presence of conditional statements within DyNN programs (\eg conditional statements). When a source program does not contain such conditional statements, a DyNN program effectively transforms into a regular DNN program, with a computational graph that can be determined statically.
Second, we recognize that DL applications typically involve pre- and post-processing stages, such as image normalization and token mapping. Consequently, the compiled DNN executable cannot function as a standalone program. Instead, a host program (\ie Listing 2) is responsible for running both the pre- and post-processing stages, as well as invoking the DNN executable for inference. It is worth noting that the host program is often developed using a high-level programming language (\eg Java, C/C++, and Python), whose compiler is equipped to adequately handle dynamism.

Based on these observations, we design \tool to move the dynamism of DyNN models to the host programs and reuse existing DL compilers to compile the sub-DNN models without dynamism.  
Specifically, \tool coverts a DyNN into multiple standard neural networks (we refer to the separated neural networks as sub-DNNs) with conditional statements determining which sub-DNNs are invoked, and then apply the existing DL compilers to compile the sub-DNNs and incorporate the conditional statements into the host program.

We identify two challenges to correctly moving the dynamism from DyNN programs to the host programs. 
The first challenge is maintaining the essential contexts (\ie concrete model instance and model input shape) for compiling each sub-DNN. To address this challenge, our insight is that each sub-DNN is a ``fixed'' part and has no dependency on the DyNN inputs. In other words, we can obtain the concrete DNN instance for each sub-DNN by performing constant propagation.
Motivated by such intuition, we developed a program rewriting engine (\secref{sec:rewrite}) that first conducts loop unrolling and then constant propagation to make each variable that has no dependency on the DyNN's input constant.
As for maintaining the input tensor shape for each sub-DNN, our insight is that each sub-DNN input is the output of its predecessor sub-DNNs.
Based on this insight, we introduced a novel concept called the Heterogeneous Control Flow Graph (HCFG) (\secref{sec:hcfg}), a special type of control flow graph, to model the dynamic behavior and the data flow of DyNNs.
After that, we propose a novel traverse algorithm (\secref{sec:compilation}) to traverse the HCFG, trace each sub-DNNs output shape, and use them as the input shapes for compiling the subsequent sub-DNNs.
By doing so, we can successfully collect the necessary context to compile each sub-DNN.

The second challenge we encountered involves co-optimizing the host program and the compiled sub-DNNs to achieve enhanced optimization. 
Our insight for tackling this challenge stems from the observation that the host program and the compiled sub-DNNs are typically executed on different devices, such as CPU and GPU. Consequently, unnecessary overhead may arise due to data transfers between these devices.
Thus, we can further optimize each sub-DNN and put the computation-free operations (\eg memory manipulation) on the host program to reduce the data transfer overheads.
We then propose two strategies to identify the computation-free operations (\eg constant assignment and tensor copy) in each sub-DNN and move the operations from the computational graph of the sub-DNN to the corresponding host program to ensure semantic equivalence.

We conducted extensive experiments to evaluate \tool. Specifically, we evaluate two open-source DL compilers, \tvm \cite{tvm} and \onnx \cite{onnxruntime}, on nine DyNN models, and we select two popular hardware platforms (\ie Nvidia TX2 and Nvidia AGX) as the backends.
The selected DyNN models are diverse in terms of model architecture, model size, and application. The selected hardware backends are popular embedded platforms for deploying neural networks to assist system decision-making.
We evaluate \tool in terms of compilation correctness and acceleration. Moreover,  we conduct an ablation study to understand the contribution of our proposed graph optimization module.
The results show that \tool can 100\% correctly compile all dynamic neural networks (\ie the final decision after the post-processing of the compiled DyNN has a 100\% consistent rate with the decision of the original DyNN model), and the maximum numeric error between the compiled DyNN and the original DyNN is around $10^{-4.72}$, significantly less than directly applying DL compiler to compile the DyNN (range from $10^0$ to $10^4$).
Moreover, the compiled executables run $1.12\times$ to $20.21\times$
faster than the original DyNNs running on the general-purpose DL frameworks, indicating the benefits of applying \tool to deploy DyNN models.
Finally, our ablation study shows that the proposed graph optimization module can further benefit the compilation process.

This paper made the following contributions.
\begin{itemize}
    \item   We conduct an empirical study to use two popular DL compilers (\ie \texttt{TVM} and \texttt{OnnxRuntime}) to compile four types of DyNNs. The study results illustrate the  limitations of the existing DL compilers when compiling DyNNs.
    
    \item We present a program rewriting approach that allows adapting many existing DL compilers to compile DyNNs correctly. The key novelty of our approach is to identify and represent the dynamism of DyNN programs in heterogeneous control flow graphs. The sub-DNNs in HCFGs are compiled individually, and our approach generates a host API to call the compiled sub-DNNs.
    
    \item Based on the novel ideas, we implement \tool; our evaluation results show \tool can correctly compile nine DyNN models and accelerate the DyNN's inference time overheads range from $1.12\times$ to $20.21\times$.
    
\end{itemize}

\section{Background}
\label{sec:background}

\subsection{Deep Learning Compiler}

\begin{figure}[t!]
    \centering
    \includegraphics[width=0.46\textwidth]{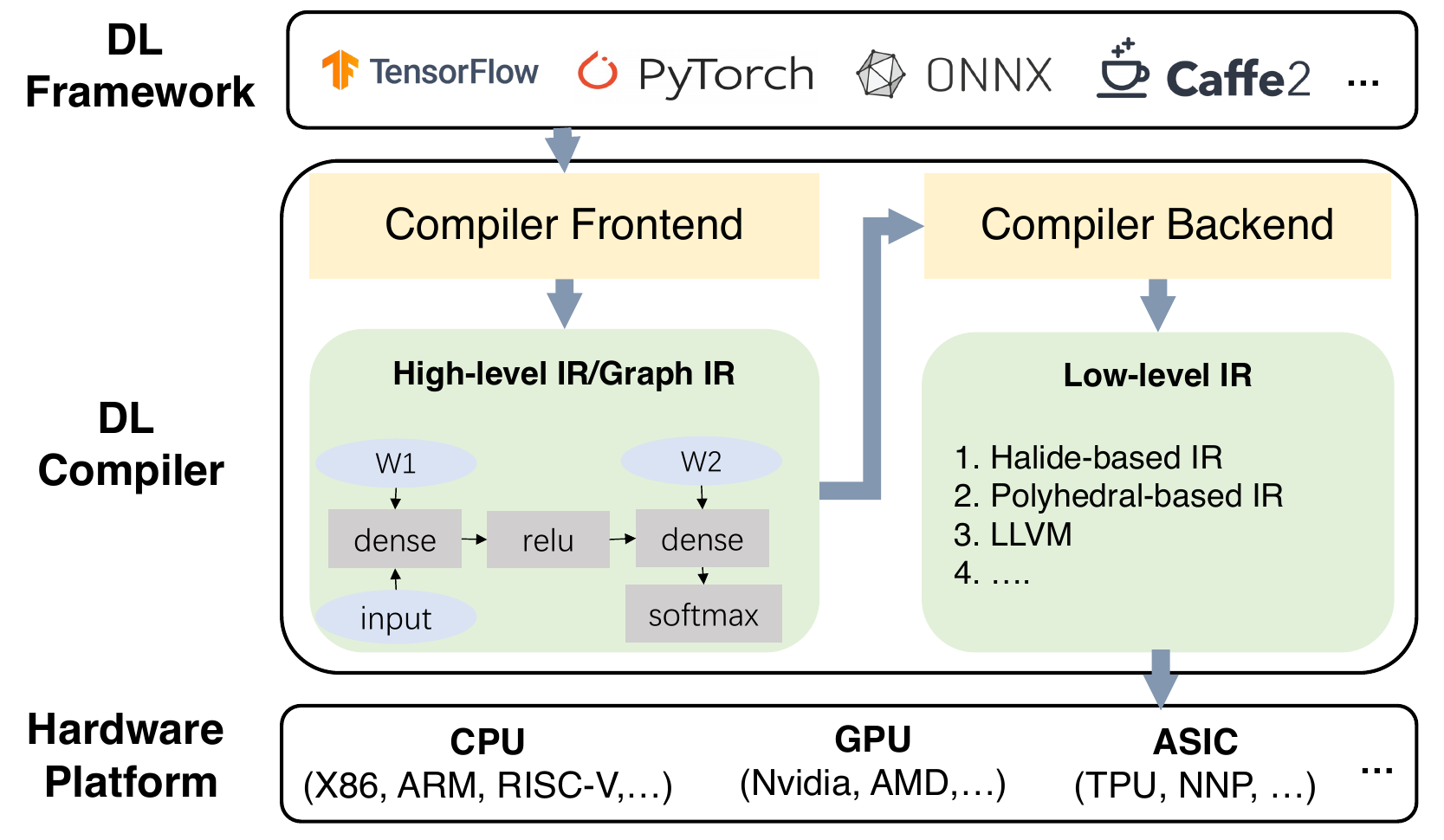}

    \caption{An overview of design architecture of DL compilers.}
    \label{fig:dlcl}
\end{figure}

DL compilers are designed to optimize  deep neural networks from high-level deep learning frameworks (\eg \texttt{Pytorch} \cite{paszke2019pytorch},  \texttt{Caffe} \cite{jia2014caffe} and \texttt{TensorFlow} \cite{li2020tensorflow}) and produce executables for AI-programs running on different hardware platforms \cite{wu2022dnd, li2020deep}.
As shown in \figref{fig:dlcl}, a DL compiler primarily contains two parts \cite{dlcompiler}: the compiler frontend and the compiler backend.
To compile a DNN, the compiler frontend first translates the DL model into high-level intermediate representations (IR) for hardware-independent optimizations.
After that, the compiler backend converts the high-level IR into low-level IR for hardware-specific optimizations and code generation.
The high-level IR in a DL compiler is typically represented by a graph, called \emph{computational graph}. In a computational graph, a node represents an operation on a tensor or a program input, and an edge represents the data dependence between operations. The low-level IRs are language and machine-dependant, capturing the hardware characteristics (\eg memory management).

\begin{lstlisting}[language=Python,label=lst:compile, caption=Example of compiling DNNs.]
# Load a pretrained model
model = torchvision.models.resnet18(pretrained=True)
model = model.eval()

# Grab the TorchScripted model via tracing
input_shape = [1, 3, 224, 224]
example_data = torch.randn(input_shape)
scripted_model = torch.jit.trace(model, example_data)

# Transfer the PyTorch model to Relay
input_name = "input0"
shape_list = [(input_name, img.shape)]
mod, params = relay.frontend.from_pytorch(scripted_model, shape_list)

# Compile the model for the target platform
target = tvm.target.Target("llvm", host="llvm")
dev = tvm.cpu(0)
with tvm.transform.PassContext(opt_level=3):
    lib = relay.build(mod, target=target, params=params)
\end{lstlisting}



\begin{lstlisting}[language=Python,label=lst:host, caption=Example host program deploying compiled DNNs.]
# Preprocess the image and convert to tensor
img = Image.open(img_path).resize((224, 224))
my_preprocess = transforms.Compose([
        transforms.Resize(256),
        transforms.CenterCrop(224),
        transforms.ToTensor(),
        transforms.Normalize(
        mean=[0.485, 0.456, 0.406], 
        std=[0.229, 0.224, 0.225]),
    ])
img = my_preprocess(img)
img = np.expand_dims(img, 0)

# Load the compiled DNN and inference
m = graph_executor.GraphModule(lib["default"](dev))
img = img.astype("float32")
m.set_input(input_name, tvm.nd.array(img))
m.run()
tvm_output = m.get_output(0)

# Postprocess the inference results
top1_tvm = np.argmax(tvm_output.numpy()[0])
tvm_class_key = class_id_to_key[top1_tvm]
\end{lstlisting}




Listing \ref{lst:compile}\footnote{\href{https://tvm.apache.org/docs/how_to/compile_models/from_pytorch.html}{https://tvm.apache.org/docs/how\_to/compile\_models/from\_pytorch.html}} shows an example using \texttt{TVM} \cite{chen2018tvm} to compile a DNN for mobile programs.
As shown in line 8, to compile a DNN, the first step is to trace the DNN. The trace step requires two inputs: a DNN instance (\ie \texttt{model}) and an input example (\ie \texttt{example\_data}). It outputs one scripted module.
Listing \ref{lst:host} shows an example of how the compiled DNNs are deployed in a mobile platform.
Lines 1-12 show the pre-processing step to normalize the input image; lines 14-17 show how to load the compiled DNN and use it for inference.
Lines 21-23 show the post-processing step.
In this paper, we refer to the program that loads the compiled DNN executable as the \emph{host program}.

\begin{figure}[bh!]
    \centering
    \includegraphics[width=0.48\textwidth]{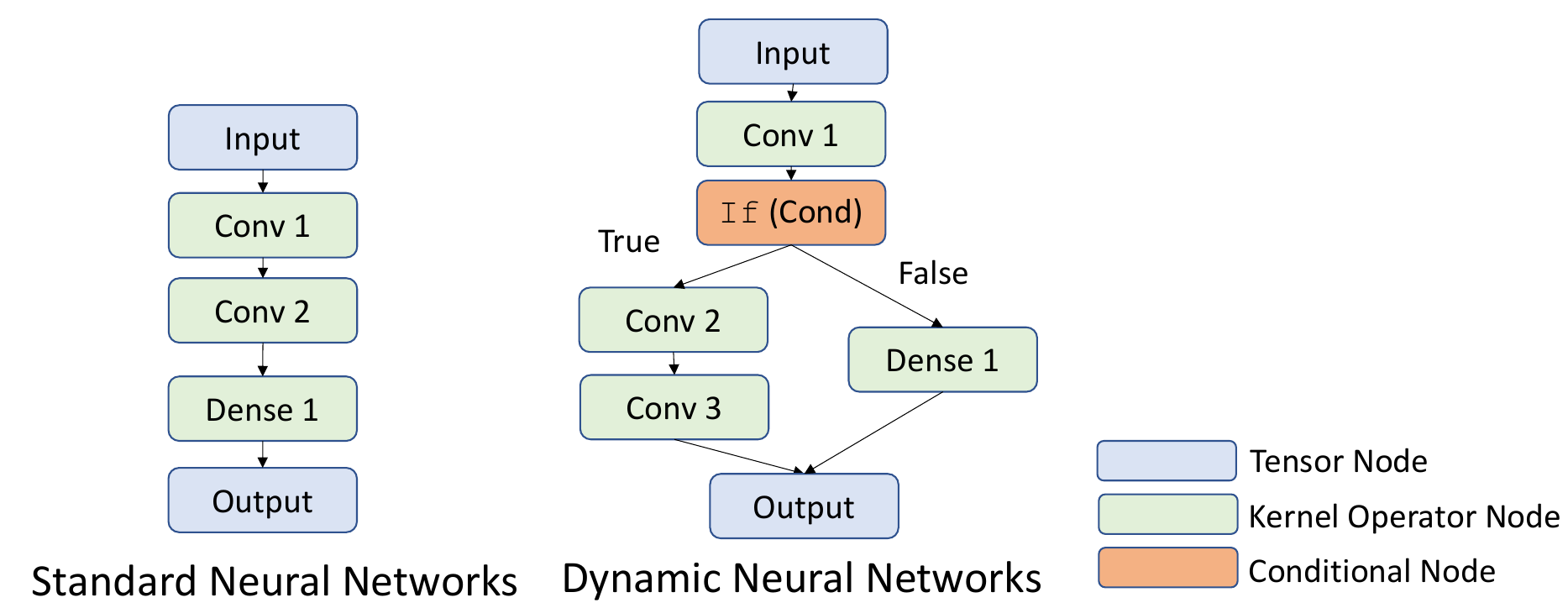}

    \caption{An overview of the standard neural networks vs. dynamic neural networks.}
    \label{fig:adnn}
\end{figure}

\subsection{Dynamic Neural Network Model}

Before we introduce Dynamic Neural Networks (DyNNs), we first introduce the basic concepts of standard neural networks. 
As shown in the left subfigure of \figref{fig:adnn}, a neural network can be abstracted as a directed acyclic graph (DAG). The node in the graph represents either the tensor or the kernel operators (\eg \texttt{convolutional} operator and \texttt{dense} operator), and the edge represents the data-flow dependency between the nodes.

However, standard neural networks cannot satisfy the modern needs of many real-world applications. Take natural language processing as an example, where the neural networks require different input and output dimensions, making dynamic neural networks inherently necessary. In response to these demands, researchers have proposed DyNNs \cite{adnn1, adnn2, adnn10, adnn9, adnn8, adnn7, adnn6, adnn5, adnn4}. DyNNs merge neural networks with conditional logic, allowing their execution paths to be modified based on varying inputs. In the right subfigure of \figref{fig:adnn}, DyNNs include additional components such as the \texttt{If} conditional nodes alongside the tensor and kernel operator nodes. When presented with specific input, the \texttt{If} conditional node utilizes the computed intermediate tensor (\eg the output tensor of the \texttt{Conv1} node in our example) to determine which sections of the overall graph should be activated for computation. For instance, in early-exit DyNNs, the \texttt{If} conditional node employs the computed confidence score from its inter-classifier to decide whether to continue the computation. Similarly, in neural machine translation DyNNs, the \texttt{If} conditional node leverages the output token value to determine whether the translation should be completed.

In this study, our focus lies on dynamic neural networks that exclusively consist of input-independent loops, as the inclusion of an input-dependent loop could potentially lead to an infinite loop scenario. DyNNs are unique tensor programs in which the parameters are learned from data rather than manually defined. However, this reliance on trained parameters renders them susceptible to adversarial examples \cite{chen2022nmtsloth}, where the adversarial agent can manipulate the inputs to deceive the DNN and create an infinite loop. Consequently, to mitigate this vulnerability, current implementations of DyNNs often incorporate a flag that enforces a constant maximum number of iterations, independent of the inputs, thus safeguarding against potential adversarial attacks.
This implementation style has been seen in existing work \cite{chen2022nmtsloth} which studied 1,455 DyNN implementations and all of the studied DyNNs apply such an implementation style.

\section{Empirical Study}
\label{sec:limitation}

In this section, we perform an empirical study to understand if existing DL compilers can correctly compile DyNN models.

\subsection{Study Setup}

\begin{table*}[htbp]
  \centering
  \caption{The Subject DyNNs in Our Preliminary Study}
    \begin{tabular}{c|clccc}
    \toprule
    \textbf{DyNNs } & \textbf{Domain} & \multicolumn{1}{c}{\textbf{Dynamic Mechanism}} & \textbf{\# of Branch} & \textbf{Github Star} & \textbf{Citation} \\
    \midrule
    Shallow-Deep   & Image Classification & layer wise early stopping DNN & 13    & 24    & 89 \\
    SkipNet & Image classification & block wise selective executation & 53    & 208   & 355 \\
    En-Decoder & Image caption & token wise caption generation & 50    & 1.2k  & 8863 \\
    AttentionNet & machine translation & token wise caption generation & 200   & 1.2k  & 40711 \\
    \bottomrule
    \end{tabular}%
  \label{tab:study_adnn}%
\end{table*}%

\fakeparagraph{Target DL Compilers}
We target two DL compilers in this study: Apache \tvm \cite{tvm} and Microsoft \onnx \cite{onnxruntime}, both of which are popular in academic research and industry work.
\tvm is a DL compiler for deploying DNNs on various platforms. It first converts a deep neural network into a static computational graph for high-level optimization and then generates hardware-specific code for each node in the transformed graph.
On the other hand, \onnx is a runtime-based framework. It also converts a deep neural network into a graph representation for optimization. After that, \onnx maps the graph node to pre-compiled kernel operator functions.

\fakeparagraph{Target DyNN Models}  We selected DyNN models for our study using the following four criteria.
The selected DyNN models
\textit{(i)} are the state-of-the-art models, which are published at top-tier conferences and outperform others according to their publication; 
\textit{(ii)} are popularly used in work conducted by both academia and industry;
\textit{(iii)} differ from each other in terms of input domains and dynamic mechanisms;
\textit{(vi)} are publicly available, and their code can be successfully executed.

Based on the above criteria, we selected four DyNNs from the survey by Huang et al. \cite{dynamicsurvey}, listed in \tabref{tab:study_adnn}.
Two of them are \emph{energy-saving DyNNs} \cite{deepperform}: Shallow-Deep and SkipNet~\cite{kaya2019shallow, wang2018skipnet}. Shallow-Deep is an early-termination DyNN that has multiple exits in a deep neural network. If one of the exits is confident about the prediction, the execution is stopped early. SkipNet is a conditional-skipping network that decides to skip or execute a DNN block based on the intermediate gate values. AttentionNet \cite{vaswani2017attention} and En-Decoder \cite{xu2015show} are \emph{generative DyNNs}. AttentionNet is a Neural Machine Translation (NMT) model that uses an attention mechanism to draw dependencies between input and output. En-Decoder uses two different types of attention mechanisms to generate captions for images.

\subsection{Study Process}

To check whether existing DL Compilers can correctly compile the DyNNs, our intuition is that a correct compilation process should not change the semantics of the DyNNs.
With this intuition, we compare the semantics of the original DyNNs and the compiled DyNNs.
Specifically, we generate random inputs and feed the generated inputs to both the original DyNN and the compiled DyNN for inference and collect the outputs.
Given the same inputs, the compiled and original DyNNs should produce identical outputs. Otherwise, the compiled DyNN is not semantically equivalent to the original DyNN, implying the compilation process fails.

Recall the original DyNN model first runs on a general-purpose DL framework (\eg \texttt{PyTorch} and \texttt{TensorFlow}) to train its parameters. 
Thus, we can use kernel functions from the general-purpose DL framework to launch the original DyNN, and we denote this program as the vendor program $\mathcal{V}(\cdot)$.
We then follow each target DL compiler's documentation to compile the original DyNN and generate the compiled executable $\mathcal{C}(\cdot)$.
We randomly sample 1,000 inputs from a DyNN model's hold-out testing dataset as the test suite.
We feed the test suite to both $\mathcal{V}(\cdot)$ and $\mathcal{C}(\cdot)$ and collect the outputs before and after the post-prepossess (\eg the variable \texttt{tvm\_output} in line 19 and \texttt{tvm\_class\_key} in line 23 in Listing~\ref{lst:host}). We use the subscripts $before$ and $post$ to distinguish these two outputs. We then define two metrics to evaluate whether the target DL compiler can produce correct programs. 
\begin{equation}
\begin{split}
    &  \delta = log \left\{  \max_{i=1}^N  ||  \mathcal{C}_{before}(x_i) - \mathcal{V}_{before}(x_i) ||   + \epsilon \right\}\\
    &  \qquad \eta = \frac{1}{N} \sum_{i=1} ^ {N} \mathbb{I}(\mathcal{V}_{post}(x_i) \neq \mathcal{C}_{post}(x_i)) 
\end{split}
\label{eq:correct_metric}
\end{equation}

\fakeparagraph{Metrics}
 As shown in \equref{eq:correct_metric}, our first metric is \textit{maximum numeric error} $\delta$, which computes the maximum numeric error between the vendor and compiled DyNN programs outputs before the post-prepossess (\ie the maximum error between the outputs of line 19 in Listing \ref{lst:host}), where $\epsilon$ is set as $10^{-10}$ to avoid the division-by-zero issue.
The DNN compilation process needs to perform some numeric matrix operations, and errors naturally exist in the matrix operations. Thus, numeric inconsistencies are inevitable in the DNN compilation process.

However, some numeric inconsistencies may not affect DyNN programs' final decision after post-processing.
Thus, besides \textit{maximum numeric error}, we propose \textit{final inconsistent rate} $\eta$ to measure the inconsistent rate of the post-prepossessed outputs between the vendor DyNN programs and the optimized DyNN programs, where $\mathbb{I}$ is the identity function, and it outputs 1 if the expression inside this function is evaluated to be true; otherwise 0.
If the \textit{final inconsistent rate} does not equal to 0, then it means that $\mathcal{C}$ will produce different outputs with $\mathcal{V}$ given the same inputs, implying the DL compiler changes the DyNN's semantics.
For classification DyNNs, the post-processing step involves computing the predicted label. This is achieved by searching for the category with the highest confidence scores \cite{he2016deep}. For generation DyNNs, we set the post-prepossess as computing the generated sequences, which is done by searching the token that has the maximum likelihood among each output position \cite{vaswani2017attention}.

\fakeparagraph{Comparison Baselines} To show the compilation process is correct, we also compile a standard DNN program with no branches, ResNet50 \cite{he2016deep}, as the baseline.

\begin{table}[tbp!]
  \centering
  \caption{The rate of inconsistency outputs predictions.}
  
  \resizebox{0.49\textwidth}{!}{
    \begin{tabular}{c|cccc|c}
    \toprule
    \textbf{DL Compiler} & \multicolumn{1}{l}{\textbf{Shallow-Deep}} & \multicolumn{1}{l}{\textbf{SkipNet}} & \multicolumn{1}{l}{\textbf{En-Decoder}} & \multicolumn{1}{l}{\textbf{AttentionNet}} & \multicolumn{1}{c}{\textbf{ResNet}} \\
    \midrule
    \textbf{OnnxRuntime} & 0.83   & 1.00  & 0.87   & 0.83    & 0.00 \\
    \textbf{TVM} & 0.83   & 1.00  & 0.87   & 0.83    & 0.00 \\

    \bottomrule
    \end{tabular}%
    }
  \label{tab:inconsistency}%
\end{table}%

\subsection{Study Results}
The number of inconsistent predictions from the compiled DL executable and the DL framework is shown in \tabref{tab:inconsistency}.
The results demonstrate that the majority of randomly selected inputs (more than 80\%) produce inconsistent outputs between vendor programs and compiled executables, implying that the produced DNN executable is semantically inequivalent to the original DyNN programs.

\begin{figure}
    \centering
    \includegraphics[width=0.33\textwidth]{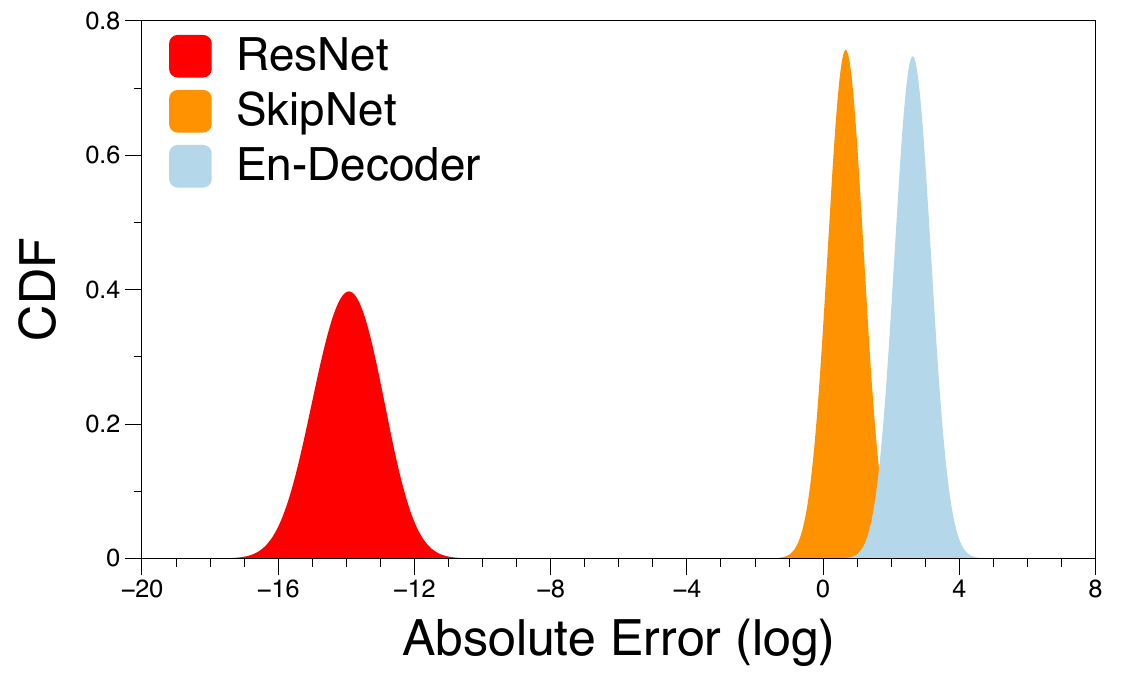}

    \caption{The error distribution of standard DNN and DyNNs.}
    \label{fig:error}
\end{figure}

\noindent\textbf{Failure Analysis.} We attempted to comprehend the enormous amount of inconsistent predictions in \tabref{tab:inconsistency}. In particular, we would like to show that the inconsistency is caused by the design limitation of the DL compiler rather than the numerical errors in the compiler optimization process.
We visualize the error distribution of the outputs from $\mathcal{V}(\cdot)$ and $\mathcal{C}(\cdot)$.
The error distribution is shown in \figref{fig:error} (for better presentation, we only show the results of SkipNet and En-Decoder, more results could be found on our website), where the x-axis represents the absolute error and the y-axis represents the cumulative distribution function (CDF) \cite{statistic}.
We observe that the absolute errors of ResNet are mostly located in the range of $[10^{-15}, \; 10^{-13}]$. However, the errors of DyNNs are located in the range of $[10^{-1}, \; 10^{4}]$, depending on the DyNN models.
Such a significant error gap (\ie more than $10^{14}\times$) demonstrates that the discrepancy in DyNN is not due to numerical errors.

To gain further insight into the reasons for compilation failure, we conducted an analysis focusing on consistent outputs (it is worth noting that some DyNNs, as shown in \tabref{tab:inconsistency}, are capable of producing consistent outputs).
In our analysis, we specifically tracked the execution traces of these inputs and compared them to the traces used during the compilation process of the DyNN. As indicated in line 8 of Listing \ref{lst:compile}, the DL compiler relies on tracing an example data to compile the model.
During our investigation, we made a crucial observation: these traces were found to be identical. This observation can be attributed to a fundamental characteristic of existing DL compilers - their ``static'' nature of the tracing mechanism in the compilation process. 
Consequently, during the compilation process, the DL compiler disregards conditional branches (such as if statements) and generates a fixed computational graph based on the execution trace of the example input. As a result, the compiled DyNN executable utilizes the same execution path for all inputs, as dictated by the fixed computational graph.
While it is possible for inputs to follow the same execution path as the example data, resulting in the compiled executable producing identical results to the original DyNN program, the likelihood of this occurring is exceptionally low. This is particularly evident in the case of SkipNet, which encompasses an astounding $2^{53}$ distinct execution paths.

\fakeparagraph{Summary} Our experimental results confirm the limitations of existing DL compilers. Specifically, while some of these compilers do offer support for control flow, they struggle to correctly compile DyNNs due to their reliance on tracing mechanisms for obtaining computational graphs.


\begin{lstlisting}[float=tp,language=Python,label=lst:syntax, caption=An example to demonstrate the challenge of identifying sub-DNNs from the source code.]

def adaptive_forward_compile(self, x):
    # some code for tensor computation
    
    for g in range(3):
        for i in range(self.num_layers[g]):
            
            if g == 0 and i < self.num_layers[g] - 1:
                i = i + 1
            
            name = 'group{}_ds{}'.format(g + 1, i)
            if name in self.attr_layers:
                model = self.attr_layers[name]
                prev = model(prev)

            if mask == 0:
                x = (1 - mask).expand_as(prev) * prev
            else:
                ly_name = 'group{}_layer{}'.format(g+1,i)
                layer = self.attr_layers[ly_name]
                x = layer(x)
                x = mask.expand_as(x) * x
                cnt = cnt + 1
            prev = x
#### Some codes 
\end{lstlisting}


\section{Challenges and Intuitions}
\label{sec:challenge}

DyNN model compilation imposes several unique challenges compared to standard ``static'' neural network model compilation.
We enumerate two major challenges and provide a high-level idea of how \tool addresses them.

\fakeparagraph{Goal: Designing a general approach for various DL Compilers using different types of IRs}
To enable existing DL compilers to effectively compile DyNNs, the initial step is to represent the dynamic behavior of DyNNs within these compilers. Furthermore, to fully leverage the capabilities of various optimization strategies present in these DL compilers, a comprehensive approach needs to be designed for each compiler, taking into account their unique designs and characteristics.

\noindent\textbf{\textit{Solution:}} 
Our solution is based  on the following two observations. First, the dynamic behavior of DyNNs comes only from conditional statements (as shown in \figref{fig:adnn}). DyNNs can be separated into multiple standard neural networks (we refer to the separated neural networks as sub-DNNs) with conditional statements determining which sub-DNN to be invoked, and each sub-DNN can be correctly compiled by existing DL compilers. Second, a compiled DL program consists of an external library function and a host program (as shown in Listing \ref{lst:host}). The host program is a program often implemented by a modern high-level programming language (\eg C/C++ and Python). These languages already have programming constructs to represent the dynamic behaviors (\ie conditional statements).
Combing these two observations, our idea is to split the conditional statements from tensor computations in DyNNs, and then apply the existing DL compilers to compile the tensor computation part and incorporate the conditional statements into the host program.

\fakeparagraph{Challenge 1: Maintaining the contexts for compiling sub-DNN} 
The first challenge is compiling each sub-DNN in the correct contexts.
Recall that in Listing \ref{lst:compile}, compiling a neural network needs two contexts: the DNN model instance and a DNN model input.
Unfortunately, both contexts are not explicitly exhibited in the DyNN implementation.
Listing \ref{lst:syntax} shows an example of a simplified implementation of SkipNet \cite{wang2018skipnet}.
From the code snippet, it is not easy to identify the necessary context for compilation because some contexts are represented in an implicit way.
For example, an existing DL compile is unable to compile a basic block (e.g., lines 13-14) because the compiler is unable to obtain the concrete model instance \texttt{model} (\ie model is determined by the variable \texttt{name} as the statement in line 13 assigns \texttt{self.attr\_layers[name]} to variable model) nor the input tensor shape of the model instance.

\noindent\textbf{\textit{Intuition 1:}} 
To address this challenge, our observation is that the dynamic behavior of the DyNN model comes from its conditional \texttt{If} node. Removing the conditional \texttt{If} node, the rest sub-DNNs are ``fixed''; thus, sub-DNNs have fixed context and have no dependency on the DyNN inputs.
In other words, we can obtain the concrete DNN instance for each sub-DNN by performing constant propagation and loop unrolling.
As for the shape of each sub-DNN's input, our intuition is that we can start from the entry of the DyNN models' computational graph, iteratively compute the output shape of each node in the graph, and set the output shape as the input shape for the successor node.

\fakeparagraph{Challenge 2: Co-optimization between the host program and tensor computation}
Recall that the compiled DyNNs will be invoked as an API function in the host program. To ensure the semantic equivalence between the generated API and the original DyNNs, for each sub-DNNs, we need to track (i) the required input variable of the corresponding code snippet and (ii) the live variable after the corresponding code snippet, and then set these variables as the input/output of the sub-DNNs.
However, the output variable of a sub-DNN may be a constant value or identical to one of the sub-DNN's input variables. Putting these variables to DL compilers to get acceleration on the accelerator (\eg GPU) may introduce unnecessary data transfer time.
Thus, it calls for a co-optimization between the host program and the tensor computation.

\noindent\textbf{\textit{Intuition 2:}} Our intuition is that DL compilers are designed to accelerate the ``computing'' for modern hardware platforms. We seek to reduce as many computation-free data transfer overheads as possible to accelerate the inference process. Based on this intuition, we propose two graph optimization strategies to further optimize the computational graph of each sub-DNN. Our optimization strategy will put computation-free data transfer left in the host program to reduce the data transfer time from the host program to the accelerator.


\begin{figure*}
    \centering
    \includegraphics[width=0.96\textwidth]{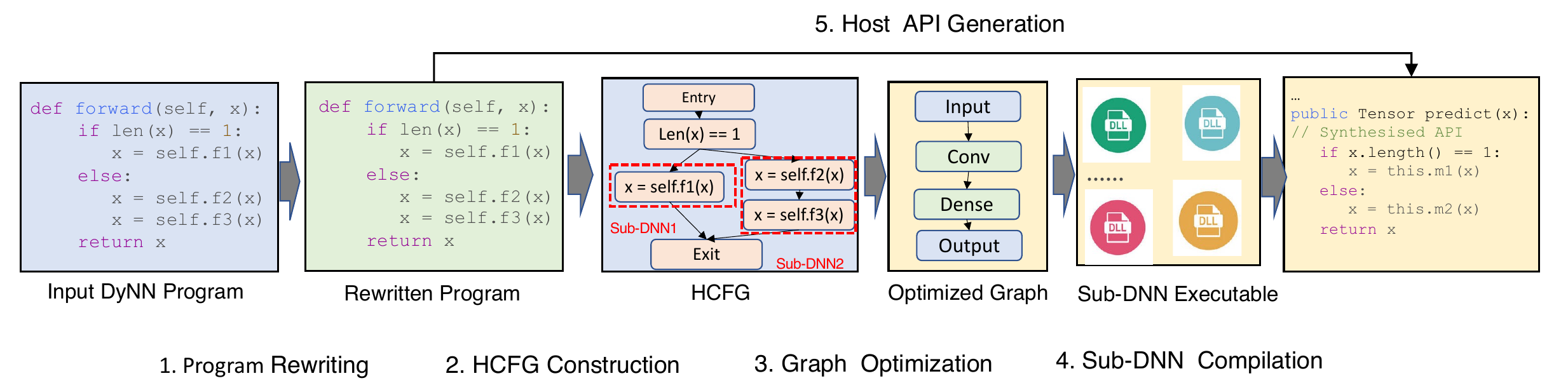}

    \caption{Design overview of \tool.}
    \label{fig:overview}
\end{figure*}

\section{Our Approach: \tool}

\label{sec:approach}

Given a DyNN model $P_{DyNN}$, our goal is to compile it and generate an optimized host API function $P_{Host}$ that is semantically equivalent to the $P_{DyNN}$. 
Formally denoted as 
\begin{equation}
   P_{DyNN}(x) = P_{Host}(x) \qquad\qquad \forall {x} \in \mathcal{X}
\end{equation}
For any inputs that belong to the program input domain, the  host API will produce the same output as the original DyNN.
Moreover, the host program can call the API using a similar way in Listing \ref{lst:host}. 
\subsection{Design Overview}

Figure \ref{fig:overview} shows the design overview of our approach.
The key insight of \tool is to partition a DyNN into several sub-DNNs without dynamism for compilation and leave the dynamism part to the host program.
Based on this insight, \tool carries out the following five steps.
\textcircled{1} \textit{DyNN source program rewriting.} Given a DyNN program from the high-level DL framework (\eg \texttt{PyTorch}), \tool first rewrites it and makes the context for each sub-DNN explicit.
\textcircled{2} \textit{HCFG Construction.} After rewriting the DyNN program, the next step is constructing a heterogeneous control flow graph (HCFG) to represent the logic conditional nodes and sub-DNN nodes.
\textcircled{3} \textit{Sub-DNN Optimization.} In this step, we build a computational graph for each sub-DNN and propose two strategies to optimize the computational graph of each sub-DNN.
\textcircled{4} \textit{Sub-DNN compilation.} After optimizing the computational graph of each sub-DNN, we start from HCFG's entry node and iteratively compile the node in the HCFG to obtain a set of compiled DNN executables. 
\textcircled{5} \textit{Host API generation.} Finally, we modify the rewritten DyNN's abstract syntax tree (AST) and covert the modified AST back to a host API function.

\begin{algorithm}[hbp!]
\caption{HCFG Construction Algorithm.} 
\label{alg:hcfg}

\begin{flushleft}
 {\bf Input:} Rewritten DyNN program $P_{DyNN}$. \\
 {\bf Output:} HCFG of the input DyNN program. \\
\end{flushleft}
\begin{algorithmic}[1]

\STATE  CFG = ConstructCFG($P_{DyNN}$)  \COMMENT{get the CFG of $P_{DyNN}$} 

\STATE  HCFG = Dict()
\COMMENT{Initlize HCFG as an dictionary}

\FOR{each N $\; \text{in} \; \text{CFG.Nodes}$}
    \STATE s = N.statements[-1] \COMMENT{get the last statement of N} 
        \IF{s is logic conditional statement}
            \STATE s, N2 = BlockPartation(N) \COMMENT{partition node N}
            \STATE HCFG.update(s) \COMMENT{Add statement s to HCFG}
            \STATE N = N2.    \COMMENT{Update N}
        \ENDIF
        \STATE HCFG.update(N) \COMMENT{Add node N to HCFG}
\ENDFOR
\end{algorithmic}
\end{algorithm}

\subsection{DyNN Program Rewriting}
\label{sec:rewrite}

Recall that the purpose of this step is to make the context for each sub-DNN explicit so that we can automatically compile each sub-DNN. To achieve this goal, we first conduct a loop unrolling process to get a program that contains no cycle in its CFG. As we introduced in \secref{sec:background}, there is no cycle in the computational graph of the DyNN model. Thus, we need to unroll all loop statements in the original DyNN program. 
Recall that our tool focuses on DyNNs that do not contain input-dependent loops. This restriction is in place to avoid the potential introduction of infinite loops, as discussed in \secref{sec:background}. Consequently, our loop unrolling technique is always feasible within this scope.
After that, we perform constant propagation to make sure that all variables have no dependency on the DyNN's input constant.
Consider the example provided in Listing \ref{lst:syntax}. Excerpts of the rewritten code are presented in Listing \ref{lst:rewrite}, showcasing the unrolling of the for loop statements from lines 4-5 and lines 9-10 of Listing \ref{lst:rewrite}. These correspond to the for loop statements found in lines 5-6 of Listing \ref{lst:syntax}. Additionally, the if statement present in lines 8-9 of Listing \ref{lst:syntax} is eliminated following constant propagation.

\begin{lstlisting}[language=Python,label=lst:rewrite, caption=The code snippet of the DyNN after rewriting]

def adaptive_forward_compile(self, x):
    # some code for tensor computation
    g = 0
    i = 1
    name = 'group1_ds1'
    #### Some repeated code 
    
    g = 0
    i = 2
    name = 'group2_ds2'
    #### Some repeated code 
\end{lstlisting}


\subsection{HCFG Construction}
\label{sec:hcfg}
After rewriting the original DyNN program, the CFG of the program will be a DAG containing no cycle, and each variable that has no data dependency on DyNN's input will be presented as a constant.

Recall that our goal is to put the dynamic logic of DyNNs on the host program.
To achieve this goal, we propose a heterogeneous control flow graph (HCFG), which is a special control flow graph (CFG). 
HCFG includes two types of nodes, the first comprises only one logic conditional statement, and the other contains multiple sequential statements.
The edges in HCFG represent the control flow paths, similar to the edges in CFG.
The HCFG construction algorithm is shown in Algorithm \ref{alg:hcfg}.
On line 1, we first construct the CFG of the rewritten DyNN program.
Then we traverse the basic blocks in the CFG (line 3), and if the last statement of the block is an if statement (line 5), we split the block into two blocks and update the HCFG (lines 6-8).
Each node in the HCFG is either an if statement or a sequence of statements that contain only the tensor computation statements.
We treat each node that contains only the tensor computation statements as a sub-DNN for compilation.

\subsection{Sub-DNN Optimization}
\label{sec:optimization}
For each sub-DNN in the constructed HCFG, we perform further optimization on its computational graph to avoid unnecessary data transfer overheads between the host program and the accelerator (\eg GPU).
As discussed in \secref{sec:challenge}, we seek to identify the computation-free operations in each sub-DNNs' computational graph and put these operations on the host program.

\begin{figure}
    \centering
    \includegraphics[width=0.48\textwidth]{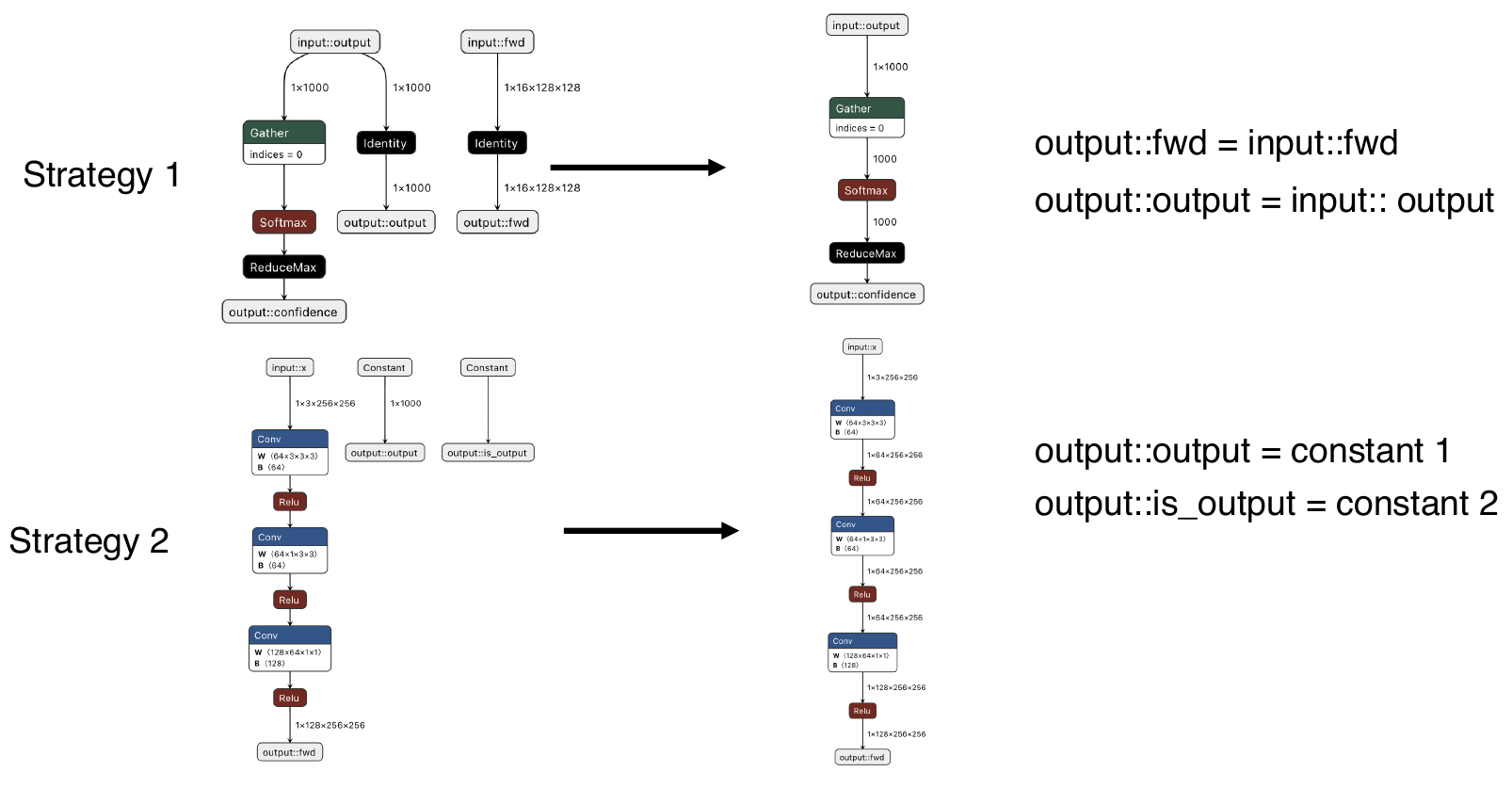}
    \caption{The proposed graph optimization strategy.}
    \label{fig:opt}
\end{figure}

The first strategy is to eliminate the identity tensor copy operation. For this strategy, we start from each output node and enumerate all paths from the input node to this output node. If a path contains only the \texttt{identity} operator, we will remove this path and add a corresponding assignment statement in the host program.
The second strategy is similar to constant propagation in compiler optimization. Firstly, we identify the output node for which all of its sink nodes (nodes without any incoming edges) are constant; we then remove these paths in the computational graph and add the corresponding assignment statement in the host program.
\figref{fig:opt} shows an example of our proposed graph optimization strategy. For each strategy, the first column shows the original computational graph, and the second column is the optimized computational graph. The third column shows the corresponding statements we add to the host program.

\subsection{Sub-DNN Compilation}
\label{sec:compilation}

\begin{algorithm}
\caption{Sub-DNN Compilation Algorithm.} 
\label{alg:compilation}

\begin{flushleft}
 {\bf Input:} Heterogeneous Control Flow Graph (HCFG).  \\
  {\bf Input:} Start Node ($N_0$).  \\
{\bf Input:} An example input of the DyNN model ($x_0$).  \\
 {\bf Output:} A mapping from node to compiled sub-DNN ($\mathcal{M}$). \\
\end{flushleft}
\begin{algorithmic}[1]

\STATE $L$ = [ $N_0$ ] \COMMENT{Maintain a search list $L$} 
\STATE $S$ = Dict() \COMMENT{Maintain $S$ to store tensor shape} 

\WHILE{$L$ is nor empty}
    \STATE $N$ = Select(L)  \COMMENT{select N based on N's precursor}
    \STATE $N_{visit}$ = True  \COMMENT{set N as visited}
    \STATE $M$ = $N.successor$ \COMMENT{collect N's successor}
    \STATE $L.update(M)$      \COMMENT{update the search list}
    \IF{$N$ is $N_0$ }
            \STATE $out =$ =\texttt{Compute($N, x_0$)} \COMMENT{compute node N}
            \STATE $S[N.id] = out$ \COMMENT{record the output shape for $N$}
            \STATE $exe =$ =\texttt{Compile($N, x$)} \COMMENT{compile node N}
            \STATE $\mathcal{M}[N.id] = exe$ \COMMENT{store the executable $N$}
        \ELSE
            \IF{$N$ is Logic Node }
                \STATE $S[N.id] = S[N.precursor]$  \COMMENT{set output shape of logic statement  the same as its precursor}
                \STATE Continue
            \ELSE
                \STATE $x$ = $S[N.precursor]$ \COMMENT{get input shape from its precursor}
                \STATE $out =$ =\texttt{Compute($N, x$)} \COMMENT{compute node N}
                \STATE $S[N.id] = out$ \COMMENT{record the output shape}
                \STATE $exe =$ =\texttt{Compile($N, x$)} \COMMENT{compile node N}
                \STATE $\mathcal{M}[N.id] = exe$ \COMMENT{store the executable}
            \ENDIF
        \ENDIF

\ENDWHILE
\end{algorithmic}
\end{algorithm}

Recall that using the DL compiler to compile a DNN model requires feeding an example input to the compiler (\ie line 8 in Listing \ref{lst:compile}). To create such an example input, the input tensor shape must be manually defined, which is a time-consuming process for DyNNs with hundreds of sub-DNNs.
To address this challenge, we propose an automatic algorithm to compile each sub-DNNs.

Our automatic sub-DNN compilation algorithm is shown in Algorithm \ref{alg:compilation}, which takes an HCFG, a start node $N_0$, and a DyNN input $x_0$ as inputs and outputs a mapping $\mathcal{M}$ from node id to the compiled sub-DNNs.
In general, the algorithm maintains a search list $L$ and a dictionary $S$ that stores the output shape for each node.
While the search list $L$ is not empty, we select a node $N$ from $L$ iteratively, depending on whether all  predecessors of $N$ have been computed (line 4).
We then set $N$ as visited and collect all non-visited successors of $N$ to update our search list (lines 5-7).
For each node $N$, there are three conditions: (1) $N$ is the start node $N_0$; (2) $N$ is the logic conditional node; (3) $N$ is a regular node other than $N_0$.
For the first condition, we compute $N$'s output shape and compile node $N$ using input $x_0$ (lines 8-12).
For the second condition, we set $N$'s output shape the same as its  predecessor's output shape because the conditional statement will not modify the variable in the program (lines 14-16).
As for the third condition, we first obtain an input $x$ from node $N$'s  predecessors, then use $x$ to compile $N$, and finally compute $N$'s output shape (lines 18-23).
Our algorithm iteratively fetches $N$ from the search list and compiles $N$ until $L$ is empty (lines 3-4).
Finally, our algorithm outputs the mapping $\mathcal{M}$, whose key is the node id and the value is the compiled sub-DNN.

\subsection{Host API Generation}

Finally, we use a template-based approach to generate the code that invokes each compiled sub-DNN. We modify the rewritten DyNN program's AST by replacing the original tensor computation statements with the statements to invoke our compiled sub-DNNs. We generate our host API function from the modified AST.
Our AST modification step only replaces the original tensor computation statements in the original DyNN program with a compiled library function call. Considering the fact that existing DL compilers have proven correct and effective in compiling standard neural networks without conditional statements, the generated API call will be semantically equivalent to the original DyNN model.

\section{Evaluation}

\subsection{Experimental Setup}

In this section, we evaluate \tool with empirical experiments and answer the following research questions. Our code and data are available on our website \cite{dycl}.\footnote{\href{https://github.com/SeekingDream/ISSTA23_DyCL}{https://github.com/DyCL}}

\begin{itemize}
    \item \textbf{\textit{RQ1 (Correctness)}:} Can \tool correctly compile DyNNs and produce semantic equivalent API for host programs?

    \item \textbf{\textit{RQ2 (Acceleration)}:} Can \tool optimize DyNNs over multiple platforms in terms of execution time?
    
    \item \textbf{\textit{RQ3 (Ablation Study)}:} 
    How much does the proposed graph optimization module in \tool enhance the execution time?
    
    \item \textbf{\textit{RQ4 (Overhead)}:} What is the overhead of \tool in compiling DyNNs?

\end{itemize}


\fakeparagraph{DyNN models} In addition to the four DyNN models in \tabref{tab:study_adnn}, we apply \tool on other five DyNN models. The IDs and names of our subject DyNNs are shown in columns 1 and 2 in \tabref{tab:correct} and we will use IDs to represent these DyNNs in all future tables. 
The selected DyNN models cover different model architectures (\eg MobileNet and ResNet), different applications (\eg image classification and text generation), and different scales (\eg model sizes range from 2.3MB to 430MB).
More detailed information about the evaluated DyNN models can be found on our website.

\fakeparagraph{DL Compilers} We consider the DL compilers used in our empirical study as the target compilers: \tvm and \onnx.
    
\fakeparagraph{Hardware Platforms} We choose two different NVIDIA platforms imposing different architectural features as our hardware platforms to show \tool can benefit the process of deploying DyNN models on various hardware platforms.
The first hardware platform is NVIDIA Jetson TX2, which has 6 ARM-based cores and a 256-core Pascal-based GPU. 
The second platform is NVIDIA Jetson AGX Xavier \cite{agx}, a powerful platform for robotics and autonomous driving with an 8-core NVIDIA Carmel CPU and a 512-core Volta-based GPU.

\noindent\textbf{Comparison Baselines.}
Recall that DyNNs compiled by existing DL compilers can only correctly infer the inputs that have the same execution paths as the example input used in the compilation process. Thus, comparing \tool with existing DL compilers is meaningless in terms of correctness.
Therefore, for the correctness (RQ1) and the acceleration (RQ2) evaluation, we compare  \tool with original DyNNs without compilation (\ie using the original DL framework). We refer to the baseline as \texttt{vendor}.

\noindent\textbf{Experimental Process and Metrics.} 
For RQ1, we use \tool to compile each \texttt{vendor} version DyNN $\mathcal{V(\cdot)}$ and get the compiled version $\mathcal{C(\cdot)}$. 
Similar to our study process in \secref{sec:limitation}, we feed 100 randomly generated inputs to these two versions and collect the outputs before and after the post-prepossess (\ie as shown in the last two lines in Listing \ref{lst:host}), and use the metrics defined in \equref{eq:correct_metric} to evaluate the correctness of compilation.
If \tool can correctly compile the DyNN model, then the results for the metric \textit{final inconsistent rate} should be zero, and the results for the metric \textit{numeric maximum  error} should be significantly different from the numeric errors in \figref{fig:error}.
       
For RQ2, we first deploy the \texttt{vendor} version DyNN $\mathcal{V(\cdot)}$ and the compiled version $\mathcal{C(\cdot)}$ on two different hardware platforms. After that, we feed the same inputs to these two versions for inference and record the inference overheads. For each input, we infer five times and report the average inference overheads.

For RQ3, we remove the graph module in \tool and apply \tool to compile each DyNN and get the compiled DyNN $\mathcal{C}_{no}(\cdot)$. After that, we deploy $\mathcal{C}_{no}(\cdot)$ and $\mathcal{C}(\cdot)$ on five different hardware platforms and evaluate their inference overheads, similar to the process in RQ2. Intuitively, if our proposed co-optimization method can benefit the compilation process, then the compiled DyNN $\mathcal{C}(\cdot)$ will have lower runtime overhead.
        
For RQ4, we report the time overheads of each module in \tool when compiling each DyNN.

\subsection{RQ1: Correctness}


\begin{table}[tbph]
  \centering
  \caption{Correctness of \tool.}
  \label{tab:correct}
  \resizebox{0.48\textwidth}{!}{
   \begin{tabular}{c|l|cc|cc|cc|cc}
    \toprule
    \multirow{3}[2]{*}{\textbf{ID}} & \multicolumn{1}{c|}{\multirow{3}[2]{*}{\textbf{Base DNN}}} & \multicolumn{4}{c|}{\textbf{Final Inconsistent Rate}} & \multicolumn{4}{c}{\textbf{Max Numeric Error}} \\
          &       & \multicolumn{2}{c|}{\textbf{Nvidia TX2}} & \multicolumn{2}{c|}{\textbf{Nvidia AGX}} & \multicolumn{2}{c|}{\textbf{Nvidia TX2}} & \multicolumn{2}{c}{\textbf{Nvidia AGX}} \\
          &       & \textbf{TVM} & \textbf{OnnxR} & \textbf{TVM} & \textbf{OnnxR} & \textbf{TVM} & \textbf{OnnxR} & \textbf{TVM} & \textbf{OnnxR} \\
    \midrule
    \textbf{1 } & \textbf{MobileNet} & 0  & 0  & 0  & 0  & -10.00  & -10.00  & -10.00  & -10.00  \\
    \textbf{2 } & \textbf{VGG19} & 0  & 0  & 0  & 0  & -9.17  & -9.35  & -9.17  & -9.35  \\
    \textbf{3 } & \textbf{ResNet50} & 0  & 0  & 0  & 0  & -4.84  & -4.94  & -4.89  & -5.53  \\
    \textbf{4 } & \textbf{WideResNet} & 0  & 0  & 0  & 0  & -4.72  & -5.60  & -4.75  & -5.48  \\
    \textbf{5 } & \textbf{ResNet38 + RNN} & 0  & 0  & 0  & 0  & -5.62  & -5.81  & -5.62  & -5.81  \\
    \textbf{6 } & \textbf{ResNet38 + Dense} & 0  & 0  & 0  & 0  & -6.45  & -6.45  & -6.45  & -6.45  \\
    \textbf{7 } & \textbf{ResxNext + LSTM} & 0  & 0  & 0  & 0  & -10.00  & -10.00  & -10.00  & -10.00  \\
    \textbf{8 } & \textbf{GoogLeNet+ LSTM} & 0  & 0  & 0  & 0  & -10.00  & -10.00  & -10.00  & -10.00  \\
    \textbf{9 } & \textbf{FlatResNet32} & 0  & 0  & 0  & 0  & -10.00  & -10.00  & -10.00  & -10.00  \\
    \bottomrule
    \end{tabular}%
    }
  \label{tab:addlabel}%
\end{table}%

The compilation correctness results are shown in \tabref{tab:correct}. Column 2 shows the names of the DyNN models under evaluation; columns 3 to 6 show the final inconsistent rate ($\eta$ in \equref{eq:correct_metric}) between the outputs $\mathcal{V}_{post}(\cdot)$ and $\mathcal{C}_{post}(\cdot)$, and the last four columns show the maximum numeric error ($\delta$ in \equref{eq:correct_metric}) between the outputs $\mathcal{V}_{before}(\cdot)$ and $\mathcal{C}_{before}(\cdot)$.

We made the following observations. 
First, for all the evaluation subjects, \tool can succesfully compile the original DyNN models for deployment, and the final outputs of the deployed host programs (\ie the results after post-process) are the same as the final outputs of the original DyNNs. 
This shows that the \tool compiled DyNN is semantic-equivalent to the original DyNN, indicating that \tool can correctly compile the DyNN model.
Second, the differences of \textit{maximum numeric error} between the outputs of the compiled DyNN and original DyNN are small, ranging from $0$ to $10^{-4.72}$. 
The differences of numeric errors are within an allowed error range (\ie \texttt{PyTorch} sets a maximum default error as $10^{-5}$ for the compilation process). 
Recall that in \figref{fig:error}, directly applying existing DL compilers to compile the DyNN models results in the maximum error range $[10^0, 10^4]$.
The small differences of maximum numeric error also confirm the correctness of \tool; otherwise, if there is a difference between the execution path between the original DyNN and the \tool compiled DyNN, the maximum numeric error will be much larger.

\begin{center}
\begin{tcolorbox}[colback=gray!10,
                  colframe=black,
                  width=0.46\textwidth,
                  arc=1mm, auto outer arc,
                  boxrule=0.95pt,
                 ]
Answers to \textbf{RQ1}: With \tool, existing ``static'' DL compilers can successful compile the DyNN models. The numeric error between the original DyNN and the compiled DyNN is minimal and does not affects the final prediction.
\end{tcolorbox}
\end{center}


\begin{figure}[hbph]
    \centering
    \includegraphics[width=0.48\textwidth]{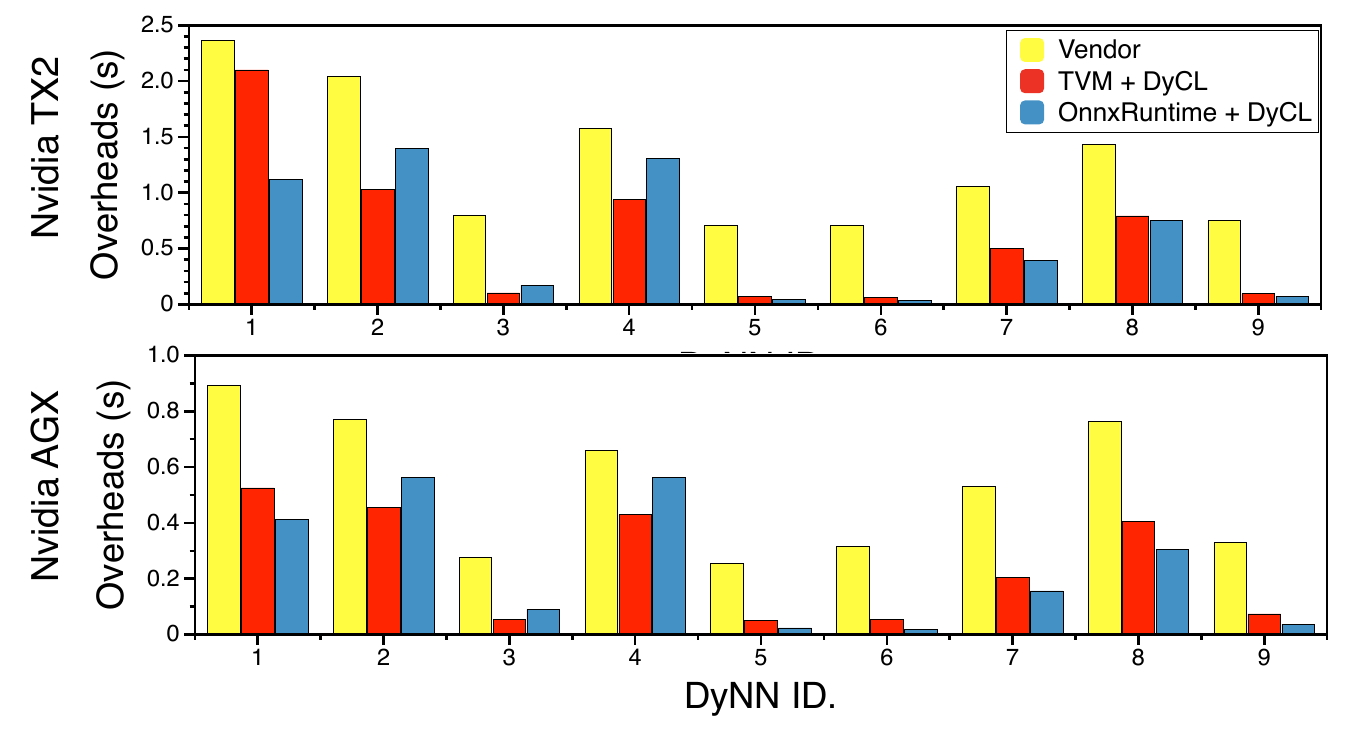}
    \caption{The inference overhead of the original DyNNs and the \tool compiled DyNNs. }
    \label{fig:accelerate}
\end{figure}

\subsection{RQ2: Accelaration}

The overheads of the original DyNN model and the compiled DyNN model are shown in \figref{fig:accelerate}, where the x-axis represents the DyNN model ID, and the y-axis shows the average inference overheads in seconds. The different color bars represent different running mechanisms.
We observe that for all our experimental settings, the compilation process can accelerate the DyNN models' inference, and the acceleration rate range from $1.12\times$ to $20.21\times$. 
Such significant acceleration shows the advantage of compiling the DyNN model for deployment, especially for deploying the DyNN model on resource-constrained platforms.
Another interesting observation is that no DL compiler can consistently outperform the others on all experimental subjects, indicating that the optimization strategies in existing DL compilers are complimentary.
This result demonstrates the benefit of \tool's capability of adapting different DL compilers to compile DyNNs.

\begin{center}
\begin{tcolorbox}[colback=gray!10,
                  colframe=black,
                  width=0.46\textwidth,
                  arc=1mm, auto outer arc,
                  boxrule=0.95pt,
                 ]
Answers to \textbf{RQ2}: \tool compiled DyNN models are consistently faster than the original DyNN models in terms of inference time, across different platforms and DL compilers. 
\end{tcolorbox}
\end{center}

\subsection{RQ3: Ablation Study}

The results of our ablation study are shown in \tabref{tab:abalition}. We show the results on Nvidia AGX platform due to the limit of space, more results could be found on our website.
Column 1 shows the same subject DyNN IDs as in \tabref{tab:correct}.
The data in columns $\mathcal{C}$ and $\mathcal{C}_{no}$ represent the inference time overheads of the programs that are compiled by \tool with and without the graph optimization module, respectively.
We observe that for most settings, the graph optimization module can accelerate the compiled DyNN with an average acceleration rate ranging from 4.007\% to 9.438\%.
Considering that existing DL compilers have almost done extreme optimization on the tensor computation on modern hardware platforms, the acceleration results in \tabref{tab:abalition} are significant.
This result confirms that our graph optimization strategy can benefit \tool in compiling dynamic neural networks.


\begin{table}[tbp!]
  \centering
  \caption{The inference overheads of compiled DyNN that are compiled with and without the graph optimization module.}

  \resizebox{0.45\textwidth}{!}{
    \begin{tabular}{c|ccc|ccc}
    \toprule
    \multirow{2}[2]{*}{\textbf{ID.}} & \multicolumn{3}{c|}{\textbf{TVM}} & \multicolumn{3}{c}{\textbf{ONNX}} \\
          & $\mathcal{C}_{no}$ & $\mathcal{C}$ & \textbf{Accelerate } & $\mathcal{C}_{no}$ & $\mathcal{C}$ & \textbf{Accelerate } \\
    \midrule
    \textbf{1 } & 0.530  & 0.525  & 0.939  & 0.417  & 0.412  & 1.222  \\
    \textbf{2 } & 0.589  & 0.455  & 29.449  & 0.621  & 0.562  & 10.576  \\
    \textbf{3 } & 0.056  & 0.055  & 1.325  & 0.088  & 0.090  & -2.524  \\
    \textbf{4 } & 0.525  & 0.429  & 22.297  & 0.585  & 0.564  & 3.685  \\
    \textbf{5 } & 0.058  & 0.051  & 13.030  & 0.026  & 0.024  & 10.241  \\
    \textbf{6 } & 0.054  & 0.053  & 1.873  & 0.020  & 0.019  & 6.796  \\
    \textbf{7 } & 0.236  & 0.205  & 15.482  & 0.162  & 0.157  & 3.593  \\
    \textbf{8 } & 0.407  & 0.406  & 0.329  & 0.311  & 0.304  & 2.252  \\
    \textbf{9 } & 0.074  & 0.074  & 0.219  & 0.037  & 0.037  & 0.223  \\
    \midrule
    \textbf{Avg} & 0.281  & 0.250  & 9.438  & 0.252  & 0.241  & 4.007  \\
    \bottomrule
    \end{tabular}%
    }
  \label{tab:abalition}%
\end{table}%

\begin{center}
\begin{tcolorbox}[colback=gray!10,
                  colframe=black,
                  width=0.46\textwidth,
                  arc=1mm, auto outer arc,
                  boxrule=0.95pt,
                 ]
Answers to \textbf{RQ3}: The graph optimization module of \tool further accelerates the compiled DyNNs, on average 4\% to 9\% faster than those compiled without the module.
\end{tcolorbox}
\end{center}

\subsection{RQ4: Overheads}

\begin{table}[htbp]
  \centering
  \caption{The overheads of \tool.}

  \resizebox{0.45\textwidth}{!}{
    \begin{tabular}{c|cc|cc|cc}
    \toprule
    \multirow{2}[2]{*}{\textbf{ID.}} & \multicolumn{2}{c|}{\textbf{DyCL (s)}} & \multicolumn{2}{c|}{\textbf{Original Overheads (s)}} & \multicolumn{2}{c}{\textbf{Extra Percentage (\%)}} \\
          & \textbf{Rewriting} & \textbf{Graph Opt} & \textbf{OnnxR } & \textbf{TVM} & \textbf{OnnxR } & \textbf{TVM} \\
    \midrule
    \textbf{1 } & 0.06  & 9.19  & 37.18  & 191.53  & 24.87  & 4.83  \\
    \textbf{2 } & 0.04  & 14.24  & 56.17  & 155.84  & 25.42  & 9.16  \\
    \textbf{3 } & 0.05  & 0.28  & 68.23  & 374.62  & 0.48  & 0.09  \\
    \textbf{4 } & 0.08  & 1.95  & 94.35  & 283.78  & 2.15  & 0.71  \\
    \textbf{5 } & 0.14  & 0.85  & 59.67  & 607.27  & 1.65  & 0.16  \\
    \textbf{6 } & 0.16  & 0.76  & 202.80  & 594.12  & 0.45  & 0.16  \\
    \textbf{7 } & 0.09  & 8.96  & 80.85  & 323.92  & 11.20  & 2.80  \\
    \textbf{8 } & 0.06  & 49.09  & 219.30  & 445.60  & 22.41  & 11.03  \\
    \textbf{9 } & 0.14  & 0.85  & 210.20  & 746.98  & 0.47  & 0.13  \\
    \textbf{Avg} & 0.091  & 9.573  & 114.307  & 413.740  & 9.900  & 3.230  \\
    \bottomrule
    \end{tabular}%
}
  \label{tab:extra}%
\end{table}%

\tabref{tab:extra} shows the overheads of \tool. 
The second and third columns show the overheads of the rewriting and the graph optimization module (overheads of other modules are ignorable), the fourth and fifth columns show the overheads of applying the existing DL compiler to compile all sub-DNNs, and the last two columns show the overhead percentage of \tool in terms of the total overheads of the compilation. 

From the results, we observe that the main overheads of \tool come from the graph optimization module. This is because each sub-DNN may be a computational graph with hundreds of nodes, thus emulating all paths from the input node to the output node is time-consuming. Moreover, the overheads of \tool are ignorable when compared with the overheads of applying the DL compiler to compile the sub-DNNs. \tool occupies only 3.23\% to 9.90\% percentage of the total overheads of compiling all sub-DNNs.

\begin{center}
\begin{tcolorbox}[colback=gray!10,
                  colframe=black,
                  width=0.46\textwidth,
                  arc=1mm, auto outer arc,
                  boxrule=0.95pt,
                 ]
Answers to \textbf{RQ4}: \tool is a lightweight approach and does not significantly increase  compilation overheads.
\end{tcolorbox}
\end{center}


\section{Threats to Validity}

Our selection of the DyNN systems, namely, ShallowDeep, SkipNet, AttentionNet, and En-Decoder, \etc,  might be a threat to the \textit{external validity} of our experimental conclusions. 
We have tried to alleviate this threat through the following efforts:
(1) The DyNNs are very popular, which can be seen through the number of citations of the works (\tabref{tab:study_adnn});
(2) the underlying DNN models are state-of-the-art models, which are used significantly;
(3) these systems differ from each other in terms of model architecture and functionality.
Therefore, our experimental conclusions should generally hold because of the diverse model subjects.
Moreover, it is important to address the issue of noisy latency measurements, as it can potentially impact the validity of our experimental conclusions. To tackle this challenge, we conducted multiple latency measurements and recorded both the average and variance values. Our results demonstrate that the variances are significantly smaller than the average value, indicating that the impact of system noise on our experiments is minimal and does not compromise the validity of our findings.

\section{Related work}

\textbf{Dynamic Neural Networks.} As discussed in Section \ref{sec:background}, Dynamic Neural Networks can be separated into two categories: Energy-saving DyNNs and Generative DyNNs. 
The Energy-saving DyNNs can be divided into two categories: Conditional-skipping DyNNs and Early-termination DyNN.
Among conditional-skipping models, 
Hua \textit{et al.} \cite{hua2019channel} and Gao \textit{et al.} \cite{gao2018dynamic} explored channel gating to determine computational blind spots for channel-specific regions unessential to classification. Liu \textit{et al.} \cite{liu2018dynamic} presented a new type of DyNN that utilizes reinforcement learning to achieve selective execution of neurons. SkipNet \cite{wang2018skipnet} used gating techniques to skip residual blocks.
On the other hand, Figurnov \textit{et al.} \cite{figurnov2017spatially} and Surat \textit{et al.} \cite{teerapittayanon2016branchynet} presented SACT and BranchyNet respectively, which are Early-termination DyNNs. SACT terminates the computation within a residual block early based on intermediate outputs, while BranchyNet uses separate exits within network for early termination.


\noindent\textbf{Deep Learning Compilers.} Deep learning Compilers have been one of the main focuses of the research community due to the requirement of flexibly deploying ML models on modern hardware platforms \cite{AMOS, ios, ROLLER, tensors, SparseTIR, rewriting, nnsmith, opt_survey}. Compilers like Apache TVM ~\cite{chen2018tvm}, Facebook’s Glow~\cite{glow}, Intel’s nGraph \cite{cyphers2018intel}, Nvidia’s TensorRT \cite{vanholder2016efficient}, Google’s XLA \cite{sabne2020xla} and Tensorflow Lite \cite{li2020tensorflow} are noteworthy compilers that are widely used to compile deep learning models.
 These compilers are fed with a Deep Learning model and generate highly optimized code as output. However, these compilers are not able to generate the correctly optimized code that is needed to represent DyNNs, as shown in Section \ref{sec:limitation}. 
 Recently, \texttt{Nimble} \cite{shen2021nimble} proposed a virtual machine (VM)-based compiler that can handle the control flow execution logic and the DNN kernels accordingly.
 However, such VM-based solution increases the DyNN's inference time overheads.
The contributions of \texttt{Nimble} and \tool are orthogonal,
because VM-based compiler can not generate model-persistence DNNs, and it is not feasible to re-design all existing non-VM based DL compilers run on VMs.

\section{Conclusion}

In this work, first, we study the 
limitation of existing DL compilers to compile Dynamic Neural Networks.
The significant inconsistent rate in our study results validates that existing DL compilers cannot handle dynamic neural networks.
Then, we propose a program rewriting approach to split the tensor computation and the conditional statements, apply the DL compiler to compile the tensor computation parts and leave the conditional statements to the host program.
Based on this idea, we propose \tool, the first tool that can reuse the existing ``static'' DL compiler in the context of dynamic neural networks.
Our evaluation of nine publicly available DyNN models shows that \tool can correctly compile DyNN models. Moreover, evaluation results show that \tool can achieve up to $20\times$ inference time acceleration.

\section*{ACKNOWLEDGMENTS}

This work was partially supported by NSF
grants CCF-2146443, CCF-2008905, CNS-2135625, CPS-2038727, CNS Career 1750263, and a Darpa Shell grant.

\bibliographystyle{ACM-Reference-Format}
\bibliography{ref}

\end{document}